\documentclass[10pt,twocolumn,letterpaper]{article}

\usepackage[pagenumbers]{cvpr}

\usepackage{amsmath,amssymb,booktabs,array,tabularx,multirow,calc}
\usepackage{needspace}
\usepackage{placeins} 
\usepackage{listings} 
\DeclareMathSizes{7.5}{7}{5}{5}
\usepackage{pifont} 
\usepackage{colortbl}

\definecolor{mygray}{gray}{0.94}

\newsavebox{\layerfoliotablebox}
\newlength{\layerfoliotablewidth}
\newlength{\layerfoliofigurewidth}

\newcommand{\LFyes}{\textcolor{green!50!black}{\ensuremath{\checkmark}}}
\newcommand{\LFpartial}{\ensuremath{\circ}}
\newcommand{\LFna}{\textcolor{red}{\ensuremath{\times}}}

\newenvironment{LFrevision}{\par\begingroup\color{black}\noindent}{\par\endgroup}

\patchcmd{\abstract}
  {\vspace*{12pt}}
  {\vspace*{5pt}}
  {}{}

\makeatletter
\renewcommand{\@maketitle}{%
  \newpage\null\vskip 12pt%
  \begin{center}%
    {\Large\bfseries\@title\par}%
    \ifx\@author\@empty
      \vspace{6pt}%
    \else
      \vspace{12pt}%
      {\large\lineskip .5em\begin{tabular}[t]{c}\@author\end{tabular}\par}%
      \vspace{10pt}%
    \fi
  \end{center}%
}
\makeatother

\definecolor{cvprblue}{rgb}{0.21,0.49,0.74}
\usepackage[breaklinks,colorlinks,allcolors=cvprblue,pdfusetitle]{hyperref}

\title{FolioWeaver: A LayerGraph-Grounded Agent for Layered Image Generation and Editing}
\author{%
Wentao~Lu\quad Shengzhe~Cui\quad Xihao~Wu\quad Dingxin~Yan\quad Yixuan~Qin\quad Pengfei~Fang\\[3pt]
Southeast University\\[2pt]
{\normalsize\texttt{\href{mailto:wentao.lu.cs@gmail.com}{wentao.lu.cs@gmail.com}}}
}
\hypersetup{pdfauthor={Wentao Lu, Shengzhe Cui, Xihao Wu, Dingxin Yan, Yixuan Qin, Pengfei Fang}}
\date{}

\begin{document}

\twocolumn[{
\maketitle
\begin{minipage}{\textwidth}
    \centering
    \captionsetup{type=figure,skip=1pt}
    \includegraphics[width=0.97\linewidth]{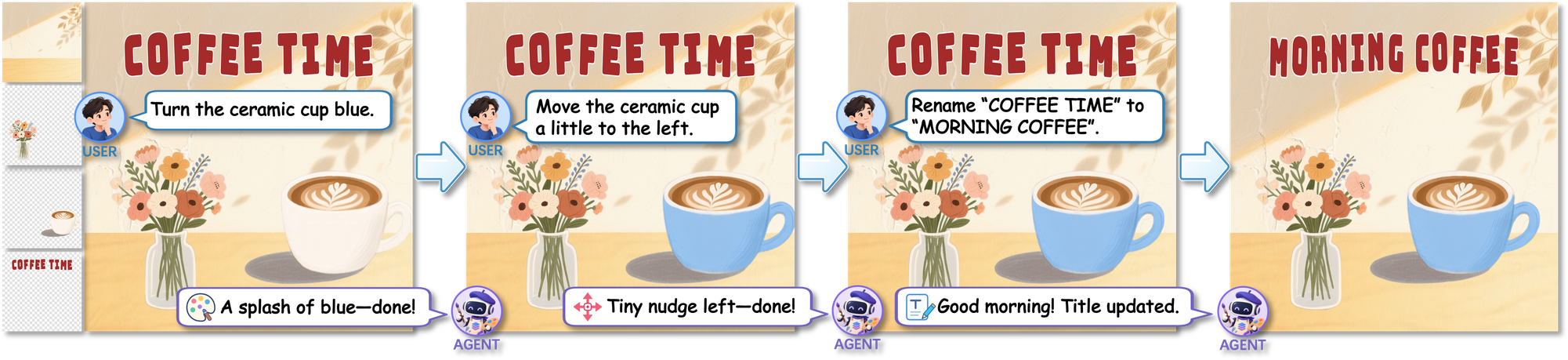}
    \caption{Illustrative example of continuous layer-level editing in FolioWeaver.}
    \label{fig:teaser}
\end{minipage}
\vspace{4pt}
}]
\begin{abstract}
Layered image generation separates images into compositable layers, allowing users to manipulate individual elements independently. However, such asset-level editability does not enable the generation system to interpret successive feedback, perform targeted edits, and preserve unaffected content. We introduce \textbf{FolioWeaver}, a LayerGraph-grounded agent that unifies layered-image generation and iterative editing through a persistent \textbf{Executable Scene State}. FolioWeaver constructs a LayerGraph and generates independent visual assets conditioned on a shared scene reference. An Editing Agent performs feedback-driven modifications and self-repair through scoped actions, selective layer execution, and visual verification, enabling targeted updates while reusing unchanged assets. We also publicly release \textbf{LayerFolio}, a dataset of \textbf{100,377} human-reviewed layered scenes spanning six visual domains, together with two evaluation protocols: \textbf{AtomicQA} for measuring requirement satisfaction and \textbf{CBLP} for assessing layer organization. Experiments demonstrate that FolioWeaver achieves superior layer organization and composite fidelity compared with direct layered-generation methods. Analyses validate the contributions of component adaptation, execution-grounded repair, and cross-domain training. FolioWeaver advances layered-image generation toward persistent, controllable, and iterative visual creation.
The dataset is available at \url{https://huggingface.co/datasets/FolioWeaver/LayerFolio}.
\end{abstract}

\section{Introduction}
\label{sec:intro}

Layered image generation maps a text prompt to a set of layers that can be composited into a complete image, rather than a single flattened output~\cite{layerdiffuse,prismlayers}. By separating content such as objects, text, and backgrounds, layered representations provide explicit units for moving, scaling, replacing, and recomposing visual elements~\cite{qwen_image_layered}. Beyond visual appearance, this task preserves the internal organization of an image, allowing users to modify selected components without regenerating the entire scene. Its value therefore lies not merely in producing multiple layers, but in providing a foundation for continued refinement after generation.

This raises a question: \textit{can a model that generates editable layers revise the same design in response to successive user feedback?} Figure~\ref{fig:teaser} illustrates this need: after generating a coffee poster, a user asks to ``make the cup blue'', ``move it to the right'', and change the title from ``Coffee Time'' to ``Morning Coffee''. These requests are not independent generation tasks: each must update the intended content while retaining unrelated components and completed edits. For a one-shot prompt-to-layer pipeline that does not maintain a scene state, the link between feedback and editing actions remains implicit. The system must re-establish the edit target and scope, and responding through global regeneration risks altering unrelated content. \textit{Producing editable layers by itself does not make the generating model capable of continued local editing.}

\textit{Our goal is to turn layered images from one-shot outputs into a persistent, executable scene state maintained by an agent, enabling scoped, verifiable updates from language feedback while preserving unaffected assets.} To this end, we introduce \textbf{FolioWeaver}, a visual creation agent built around \textbf{LayerGraph}. LayerGraph binds layer identities, semantics, appearance, layout, and inter-layer relations to concrete pixel assets, providing a shared scene representation for initial generation and subsequent editing. Rather than serving only as a layout plan before synthesis, it remains an operational interface throughout the creative process: it records the current composition and grounds new feedback in specific layers and editable attributes. Successive edits can therefore build on the current design rather than reconstructing the scene from scratch.

At a high level, FolioWeaver first translates a text request into a structured scene and generates a shared visual reference to guide the creation of its constituent elements. These elements are then assembled with the background and editable text into a layered composition. During subsequent interaction, the agent interprets each new instruction in the context of the maintained scene state, updates only the relevant content, and reuses unaffected assets. Each candidate edit is rendered and checked before being accepted, allowing automatic self-repair and user-directed editing to follow the same process. We train the generation components with paired supervision and optimize the editing policy using supervised fine-tuning and execution-grounded GRPO~\cite{deepseekmath}.

To support this workflow, we introduce \textbf{LayerFolio}, a dataset of \textbf{100,377 manually reviewed scenes} spanning six visual domains and 55 subcategories. Each scene pairs its description and relational LayerGraph with a composite image and RGBA layer assets, providing aligned supervision for scene planning and reference-conditioned object generation. We also introduce two complementary evaluation protocols: building on question-based faithfulness evaluation~\cite{tifa,dsg}, \textbf{AtomicQA} measures whether the composite satisfies prompt-level requirements, while \textbf{CBLP} assesses target-category coverage and cross-category isolation in delivered layers. Together, these resources support the training and evaluation of both scene content and its layer organization.
Our contributions are summarized as follows:
\begin{itemize}
    \item We introduce \textbf{FolioWeaver}, a LayerGraph-grounded agent that unifies layered-image generation and feedback-driven editing through a persistent Executable Scene State, with the policy optimized from executed outcomes.
    \item We publicly release \textbf{LayerFolio}, a human-reviewed dataset of 100,377 layered scenes across six domains and 55 subcategories, and introduce \textbf{AtomicQA} and \textbf{CBLP} for evaluating prompt adherence and layer organization.
    \item Extensive experiments demonstrate that FolioWeaver achieves the best overall generation performance among direct layered-generation methods while supporting continuous, feedback-driven layer editing.
\end{itemize}

\section{Related Work}
\label{sec:related}

\subsection{Layered Image Generation and Decomposition}
LayerDiffuse~\cite{layerdiffuse} introduces latent transparency for single- and multi-layer synthesis, while ART+ uses PrismLayers~\cite{prismlayers} supervision to train a multi-layer image generator. Qwen-Image-Layered~\cite{qwen_image_layered} decomposes RGB images into editable RGBA layers, and UniWorld-Design~\cite{uniworld} supports instruction-conditioned decomposition and targeted layer extraction. These approaches provide complementary routes to producing manipulable layers~\cite{designedit}. Our approach focuses on connecting layer synthesis to subsequent editing: a shared scene reference conditions object generation, while LayerGraph binds the assets to semantic, spatial, and relational state. This representation serves both initial scene construction and subsequent feedback-driven revision.

\subsection{Agents for Visual Generation and Editing}
GenArtist~\cite{genartist} uses a multimodal agent to plan tool invocations, verify outputs, and perform self-correction. LayerCraft~\cite{layercraft} combines agent coordination, dependency-aware layout reasoning, and reference-guided object integration for iterative generation and editing. GenAgent~\cite{genagent} further trains multi-turn generation and reflection through supervised fine-tuning and reinforcement learning. FolioWeaver emphasizes scoped actions over a persistent, executable layer state. Its actor inspects rendered results, applies authorized LayerGraph edits, and invokes incremental execution to reuse assets whose dependencies remain unchanged. The repair policy is learned from the visual consequences of executed actions, connecting policy optimization to the actual outcomes of layer-level modifications.

\subsection{Layered Datasets and Structured Supervision}
MuLAn~\cite{mulan} provides instance-level RGBA decompositions and occlusion information, while PrismLayers supplies synthetic multi-layer images with scene- and layer-level captions. OmniPSD~\cite{omnipsd} provides PSD-derived layers with structured descriptions, and LICA~\cite{lica} represents designs through typed component hierarchies and geometric and typographic metadata. UniWorld-Design organizes PSD content into semantic layer trees for decomposition and extraction supervision. LayerFolio combines existing resources with new scenes under a common schema spanning graphic design, user interfaces, 3D renders, photographic imagery, illustrations, and paintings. By associating layer semantics, appearance, geometry, stacking order, and explicit inter-layer relations with compositable pixel assets, it provides paired supervision for structural planning, reference-conditioned object generation, and composition.

\section{FolioWeaver: A LayerGraph-Based Agent}
\label{sec:method}

\begin{figure*}[htbp]
\centering
\includegraphics[width=\textwidth]{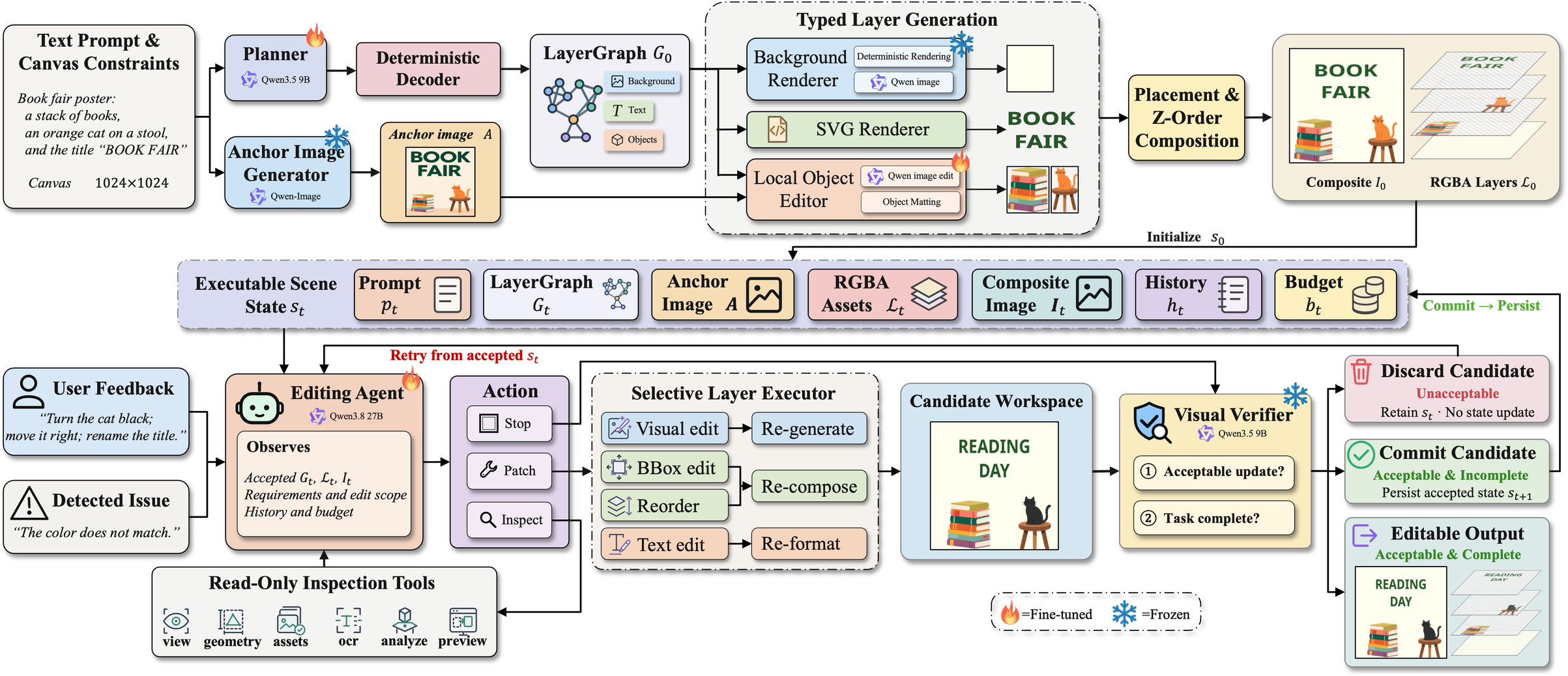}
\caption{FolioWeaver overview.
Initial generation establishes an Executable Scene State
for subsequent editing and self-repair.}
\label{fig:folioweaver_overview}
\par\vspace{-6pt} 
\end{figure*}

FolioWeaver generates an editable layered scene from a text request and
maintains it for feedback and self-repair. As shown in
Figure~\ref{fig:folioweaver_overview}, both phases share an
\emph{Executable Scene State}: the Editing Agent proposes scoped changes,
the Selective Layer Executor updates the affected layers, and the Visual
Verifier determines whether to accept the state.

\subsection{Executable Scene State}
\label{sec:layergraph_state}
A layer stack exposes editable assets but does not directly associate
language references with those assets.
We represent the scene using a LayerGraph \(G_t=(V_t,E_t)\), where \(V_t\) contains typed layer nodes and \(E_t\) encodes their relations at editing step \(t\). Each node stores an ID, description, appearance, normalized bounding box, stacking index, and asset reference; text nodes also carry editable text specifications. Directed
edges encode inter-layer relations (e.g., \texttt{supported\_by}). The graph additionally maintains canvas and scene attributes as global metadata, connecting symbolic intent to compositable pixels.

The complete \emph{Executable Scene State} at step $t$ is
\begin{equation}
  s_t=(p_t,G_t,\boldsymbol{A},\mathcal L_t,\boldsymbol{I}_t,h_t,b_t),
  \label{eq:agent_state}
\end{equation}
where $p_t$ contains the original request and active feedback, $G_t$ the
LayerGraph, $\boldsymbol{A}$ the fixed Anchor Image, $\mathcal L_t$ the RGBA assets
and caches, $\boldsymbol{I}_t$ the Composite Image, $h_t$ the interaction history,
and $b_t$ the remaining budget. The executor maintains asset references and
versions; the Editing Agent may patch only authorized content, layout,
text, and stacking fields, not layer IDs, paths, or the original request.
Explicit user feedback may override an earlier requirement.

\subsection{Initializing the Layered Scene}
\label{sec:layered_initialization}

\noindent\textbf{Text-to-Layer Planning.}
To make later edits addressable, a Planner adapted from
Qwen3.5-9B~\cite{qwen35} predicts a compact \emph{PlannerGraph} from the
prompt and optional canvas constraints. It specifies layer semantics,
boxes, order, text, and relations. A \emph{Deterministic Decoder} checks
these predictions, normalizes boxes, and supplies execution-only fields
such as asset references and relation IDs to obtain $G_0$. Thus, the
model plans content while the program handles bookkeeping.

\noindent\textbf{Reference-Conditioned Object Generation.}
Generating objects independently can disrupt visual consistency;
reusing the full prompt for every layer can introduce unwanted objects.
We instead generate one shared \emph{Anchor Image} $\boldsymbol{A}$ using the
\emph{Anchor Image Generator} (Qwen-Image~\cite{qwenimage}). For target
layer $j$, we compile a target-scoped instruction $u_j$ from its identity,
appearance, and relevant global style, excluding sibling-object requests.
The \emph{Local Object Editor}, adapted from native
Qwen-Image-Edit~\cite{qwenimage}, takes $(\boldsymbol{A},u_j)$ to produce an isolated
RGB object; RMBG-2.0~\cite{birefnet} estimates its alpha matte to form
RGBA. The target box determines the
local canvas aspect ratio, not a spatial control tensor; final placement
is handled separately.

\noindent\textbf{Typed Rendering and Composition.}
\emph{Typed Layer Generation} dispatches backgrounds to the
\emph{Background Renderer} (fill or image synthesis), objects to the
Local Object Editor, and text to the deterministic
\emph{SVG Renderer}. For each object, alpha-tight cropping and isotropic
scaling match the cropped box \emph{area} to the target box before center
alignment. The
\emph{Placement \& Z-Order Composition} module composites layers by their
stacking indices and saves the assets, placement metadata, and
image for subsequent editing.

\subsection{Feedback-Driven Editing and Self-Repair}
\label{sec:feedback_repair}

Automatic correction of an unmet initial requirement and new user
feedback enter the same loop; only the active acceptance criteria differ.

\noindent\textbf{Visual Observation and Tool-Based Inspection.}
A structurally valid graph need not produce a correct image. The
\emph{Editing Agent} observes $G_t$, $\boldsymbol{I}_t$, the active requirements,
permitted scope, and history, and may use \emph{Read-Only Inspection
Tools} for viewing, geometry, asset checks, OCR, support analysis, and
layout previews. It selects \texttt{inspect}, \texttt{patch}, or
\texttt{stop}: inspection returns evidence without modifying the accepted
graph or rendered assets;
patching proposes an authorized update; stopping requests a completion
check on the current accepted state. Neither tool output nor a stop
request commits a change.

\noindent\textbf{Scoped Actions and Incremental Execution.}
The \emph{Selective Layer Executor} validates patch schema, edit scope,
and parent version before execution. It tracks each layer's effective
generation dependencies (instruction, reference, model, dimensions,
seed, and sampling settings) to avoid unnecessary regeneration.
Bounding-box and reorder edits reuse cached RGBA assets; text edits
re-render the text layer; visual edits or explicit redraws regenerate
only the target object and matte. \emph{Unchanged assets} are reused,
although composite pixels can change due to movement or occlusion.

\noindent\textbf{Verification and State Updates.}
Each patch is executed in the \emph{Candidate Workspace}. An independent
\emph{Visual Verifier} (Qwen3.5-9B~\cite{qwen35}) checks the rendered
result against the previous state and active requirements, separately
determining whether it is \emph{acceptable} and \emph{complete}.
Candidates start from the same accepted state: rejected changes leave
$s_t$ intact, while acceptable but incomplete ones commit $s_{t+1}$.
A \texttt{stop} action undergoes the same completion check without a
patch. Upon completion, the system returns the composite, layers, and
Graph as \emph{Editable Output}; on budget exhaustion, it returns the
last accepted state.

\subsection{Learning from Supervision and Execution}
\label{sec:learning_executable}

Training comprises three stages rather than joint end-to-end
backpropagation. 

\noindent\textbf{\ding{182} Planner.} Response-only SFT adapts
Qwen3.5-9B to predict PlannerGraphs from prompts. 

\noindent\textbf{\ding{183} Local Object Editor.} We freeze native Edit weights and train
attention LoRA~\cite{lora} on full-scene reference, target instruction,
and an object image deterministically formatted from a LayerFolio RGBA
asset. For clean latent $\boldsymbol{z}_0$, noise $\boldsymbol{\epsilon}$, and $\tau\in[0,1)$, we construct \(\boldsymbol{z}_\tau=(1-\tau)\boldsymbol{z}_0+\tau\boldsymbol{\epsilon}\) and train the model to predict the target velocity \(\boldsymbol{v}^*=\boldsymbol{\epsilon}-\boldsymbol{z}_0\)~\cite{liu2023rectifiedflow}, denoting its prediction by \(\hat{\boldsymbol{v}}\).
Let \(R\in\{\mathrm{fg},\mathrm{bg},\mathrm{edge}\}\) denote the foreground, background, and boundary regions, respectively, with corresponding alpha-derived masks \(\boldsymbol M_R\). The training loss is
\begin{equation}
\begin{aligned}
  \mathcal L_R&=\frac{\sum \boldsymbol{M}_R\odot(\hat{\boldsymbol{v}}-\boldsymbol{v}^*)^2}{\sum \boldsymbol{M}_R},\\[-1pt]
  \mathcal L_{\mathrm{obj}}&=\mathcal L_{\mathrm{fg}}
  +0.25\mathcal L_{\mathrm{bg}}+0.25\mathcal L_{\mathrm{edge}}.
\end{aligned}
\label{eq:object_region_loss}
\end{equation}

The masks weight training loss only and are not inference conditions. For an empty region, we set the corresponding loss to zero.

\noindent\textbf{\ding{184} Editing Agent.} Starting from Qwen3.8-27B~\cite{qwen38_27b}, short SFT
teaches structured editing actions. Online GRPO samples
three edit trajectories for each real generated scene, executes their
proposed actions, and evaluates before/after images. Let
$Q_{\mathrm{old}}$ and $Q_{\mathrm{new}}$ be the training visual judge's
mean scores over contact, surface placement, perspective, composition,
and intent preservation, and set
$\Delta Q=Q_{\mathrm{new}}-Q_{\mathrm{old}}$. The binary degradation
indicator $d\in\{0,1\}$ equals $1$ when the mean or any criterion
falls by more than $0.08$, and $0$ otherwise. We bound the
execution-based reward as
\begin{equation}
  r=\min\!\bigl(2,\max\!\bigl(-0.95,2Q_{\mathrm{new}}+0.75\Delta Q-0.75d\bigr)\bigr).
\label{eq:repair_reward}
\end{equation}

The training judge is separate from the inference-time Visual Verifier.
GRPO~\cite{deepseekmath} uses relative rewards to update only the
Editing Agent's LoRA, freezing the remaining modules. The policy loss
covers sampled agent decisions, not tool-returned observations, thereby
optimizing actual visual outcomes rather than self-reported success.

\section{LayerFolio and Evaluation Protocols}
\label{sec:dataset}

Layered scene generation requires supervision linking scene-level requirements
with editable visual assets. We publicly release \textbf{LayerFolio}, a human-reviewed,
cross-domain dataset aligning scene descriptions, relational graphs,
composites, and RGBA layers. We also propose \textbf{AtomicQA} for prompt
satisfaction and \textbf{CBLP} for category coverage and layer isolation.

\subsection{LayerFolio: Paired Layered Supervision}
\label{sec:layerfolio}

LayerFolio is a publicly available dataset comprising \textbf{100,377 manually reviewed scenes}. Table~\ref{tab:layerfolio_comparison} compares its layer-level text,
layout, semantic-relation, and scene-taxonomy annotations with those
of existing datasets.

\begin{table*}[htbp]
   \centering
   \small
   \setlength{\tabcolsep}{3pt}
   \renewcommand{\arraystretch}{1.10}
   \sbox{\layerfoliotablebox}{%
   \begin{tabular}{@{}lcccccccc@{}}
       \toprule
       \multirow{3}{*}{\textbf{Dataset}}
       & \multirow{3}{*}{\textbf{Scale}}
       & \multirow{3}{*}{\textbf{\shortstack{\# Layers\\(range)}}}
       & \multirow{3}{*}{\textbf{\shortstack{Open\\data}}}
       & \multicolumn{4}{c}{\textbf{Annotations}} & \textbf{Curation} \\
       \cmidrule(lr){5-8}\cmidrule(l){9-9}
       & & & & \textbf{Layer} & \multirow{2}{*}{\textbf{Layout}}
       & \textbf{Semantic} & \textbf{Scene} & \textbf{Human} \\
       & & & & \textbf{text} & & \textbf{relations} & \textbf{taxonomy} & \textbf{review} \\
       \midrule
       MAGICK~\cite{magick}
       & $\sim$150K & 1
       & \LFyes & \LFyes & \LFna & \LFna & \LFna & \LFna \\
       MLCID~\cite{layerdiff}
       & $\sim$2M & 2--4
       & \LFpartial & \LFyes & \LFyes & \LFna & \LFna & \LFna \\
       Crello~\cite{crello}
       & $\sim$23K & 2--50
       & \LFyes & \LFna & \LFyes & \LFna & \LFyes & \LFna \\
       MuLAn~\cite{mulan}
       & 44,860 & 2--6
       & \LFyes & \LFyes & \LFyes & \LFna & \LFna & \LFpartial \\
       PrismLayersPro~\cite{prismlayers}
       & $\sim$20K & 2--50
       & \LFyes & \LFyes & \LFyes & \LFna & \LFna & \LFyes \\
       OmniPSD~\cite{omnipsd}
       & $>$200K & -
       & \LFpartial & \LFyes & \LFyes & \LFna & \LFna & \LFna \\
       LICA~\cite{lica}
       & $\sim$1.55M & -
       & \LFpartial & \LFyes & \LFyes & \LFna & \LFyes & \LFyes \\
       UniWorld-Design~\cite{uniworld}
       & - & -
       & \LFna & \LFyes & \LFyes & \LFna & \LFna & \LFna \\
       \midrule
       LayerFolio (Ours)
       & $\sim$100K & 2-50
       & \LFyes
       & \LFyes & \LFyes & \LFyes & \LFyes & \LFyes \\
       \bottomrule
   \end{tabular}%
   }
   \setlength{\layerfoliofigurewidth}{0.98\textwidth *
       \ratio{\dimexpr\ht\layerfoliotablebox+\dp\layerfoliotablebox\relax}
             {\dimexpr\wd\layerfoliotablebox+\ht\layerfoliotablebox+\dp\layerfoliotablebox\relax}}
   \setlength{\layerfoliotablewidth}{0.98\textwidth-\layerfoliofigurewidth}
   \begin{minipage}[t]{\layerfoliotablewidth}
   \vspace{0pt}
   \centering
   \captionsetup{position=top,skip=6pt}
   \caption{Comparison of representative layered datasets and
   transparent-asset resources.}
   \label{tab:layerfolio_comparison}
   \resizebox{\linewidth}{!}{\usebox{\layerfoliotablebox}}\par
   \end{minipage}\hfill
   \begin{minipage}[t]{\layerfoliofigurewidth}
   \vspace{0pt}
   \centering
   \includegraphics[width=\linewidth]{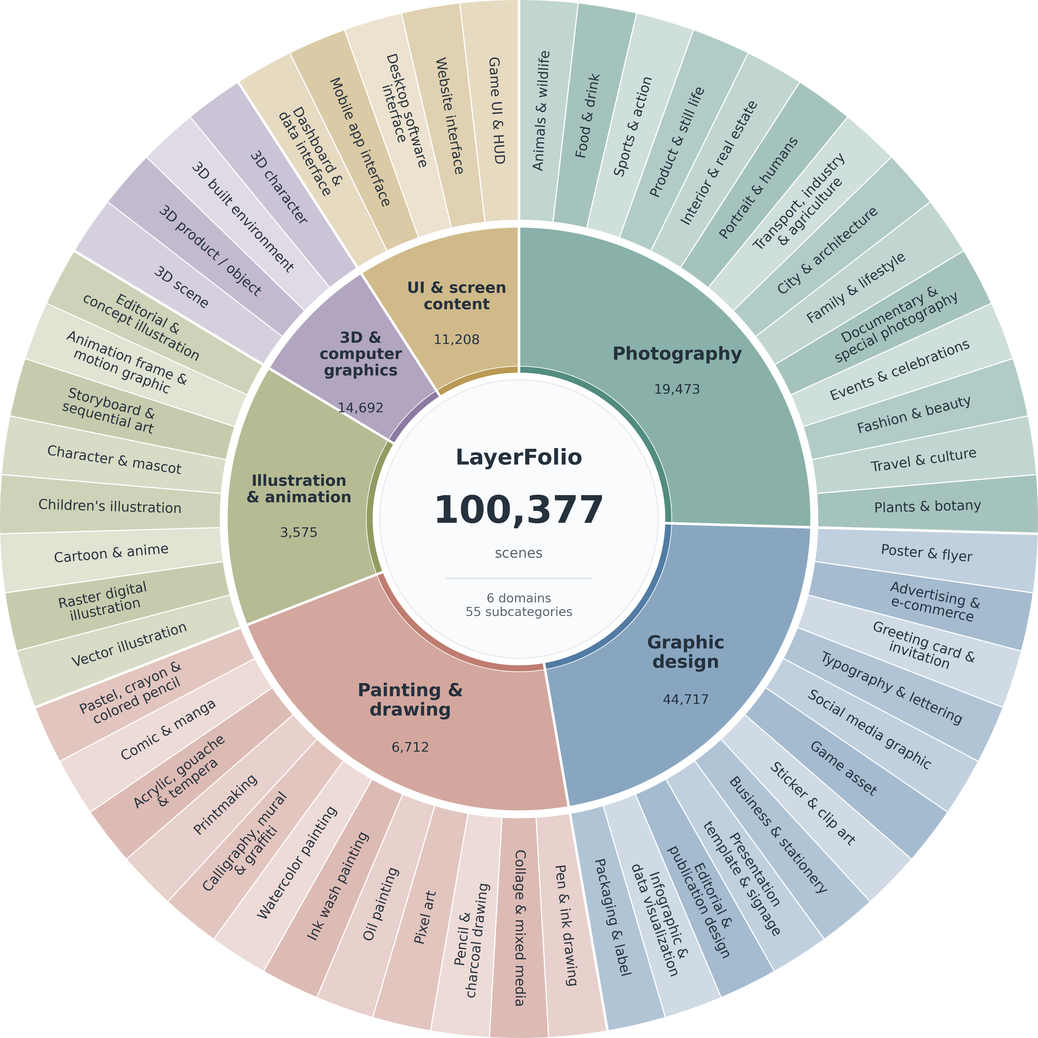}\par
   \captionsetup{type=figure,position=bottom,skip=6pt}
   \caption{LayerFolio taxonomy.}
   \label{fig:layerfolio_taxonomy}
   \end{minipage}
\par\vspace{-6pt} 
\end{table*}

\noindent\textbf{Semantic relations linked to pixel-grounded layers.}
Existing datasets provide layer descriptions, layouts, occlusion ordering,
or component hierarchies~\cite{mulan,prismlayers,omnipsd,lica,uniworld}.
LayerFolio represents each scene as a LayerGraph whose nodes link RGBA assets
to descriptions, appearance, geometry, stacking order, and optional text
specifications. Typed edges, including \texttt{supported\_by},
\texttt{occludes}, \texttt{part\_of}, and \texttt{grouped\_with}, encode
explicit inter-layer constraints.

\noindent\textbf{Cross-domain coverage with a shared representation.}
Figure~\ref{fig:layerfolio_taxonomy} summarizes LayerFolio's taxonomy
across \textbf{six visual domains and 55 subcategories}, covering graphic
design, photography, UI and screen content, 3D and computer graphics,
painting and drawing, and illustration and animation.
A unified layer schema supports entities, object groups, design elements,
and UI modules, enabling stratified sampling and domain-wise evaluation.

\noindent\textbf{Paired supervision across learning stages.}
Aligned descriptions, graphs, composites, and assets support planning,
reference-conditioned object generation, and composition. Object-generation
targets are derived deterministically from recorded RGBA assets rather than
generated anew. Holding graphs or assets fixed also enables stage-wise
diagnosis while controlling intermediate variations.

\noindent\textbf{Construction and full-dataset human review.}
LayerFolio integrates seven provenance-tracked source streams covering
existing templates and layered images, real-image-prompted synthesis,
and browser-, 2D-, and 3D-rendered scenes. Layer descriptions use isolated
RGBA evidence, relations use scene context, and geometry and asset references
are computed programmatically. Automated schema, alpha, and recomposition
checks are complemented by \emph{manual review of every retained scene}.

\subsection{Requirement and Layer Evaluation}
\label{sec:requirement_layer_evaluation}

AtomicQA measures prompt satisfaction in final composites, whereas CBLP
evaluates requested-category coverage and cross-category isolation in the
delivered layers. Both use frozen, prompt-derived criteria without ground-truth
images or method-specific graph annotations, enabling comparisons across
systems with different internal representations and layer counts.

\subsubsection{AtomicQA: Requirement Satisfaction}
\label{sec:atomicqa}

Building on question-based and likelihood-based faithfulness
metrics~\cite{tifa,dsg,vqascore}, AtomicQA converts prompts into binary
questions covering existence, attributes, counts, relations, text, and style.
\texttt{Qwen3.8-27B}~\cite{qwen38_27b} generates candidate questions, while
\texttt{Gemma-4-26B-A4B-it}~\cite{gemma4} audits their prompt grounding, coverage,
redundancy, and object binding. Each retained question tests one requirement,
with \emph{Yes} consistently indicating satisfaction, including for explicit
negative constraints. The audited question bank is frozen before image
evaluation and reused across methods.

Two frozen vision-language judges, \texttt{Qwen3-VL-\allowbreak8B-Instruct}~\cite{qwen3vl} ($m=1$) and
\texttt{Llama-3.2-\allowbreak11B-Vision-Instruct}~\cite{llama} ($m=2$), score each question. For scene
$i$, question $Q_{ik}$, and composite $I_i$, let
$\lambda_{ik}^{(m)}(a)$ denote judge $m$'s conditional log-likelihood of
answer sequence $a\in\{\mathrm{Yes},\mathrm{No}\}$. Averaging the judges'
normalized Yes supports gives
\begin{equation}
    p_{ik}=\frac{1}{2}\sum_{m=1}^{2}\sigma\!\left(
      \lambda_{ik}^{(m)}(\mathrm{Yes})-
      \lambda_{ik}^{(m)}(\mathrm{No})\right),
    \label{eq:atomicqa_support_models}
\end{equation}
where $\sigma(x)=(1+e^{-x})^{-1}$. The averaging occurs at the question
level, before scene-level aggregation.

For a scene with $M_i>0$ retained questions, the arithmetic and geometric
means are
\begin{align}
    \mathrm{AM}_i
    &=\frac{1}{M_i}\sum_{k=1}^{M_i}p_{ik},
    \label{eq:atomicqa_am}\\
    \mathrm{GM}_i
    &=\exp\!\left[\frac{1}{M_i}\sum_{k=1}^{M_i}
      \log\max(p_{ik},10^{-8})\right].
    \label{eq:atomicqa_gm}
\end{align}

The final AtomicQA-AM and AtomicQA-GM scores average the corresponding
scene-level scores over the $N$ evaluated scenes, weighting scenes equally
regardless of question count. AM captures mean requirement support, whereas
GM emphasizes weakly supported requirements. Neither is a calibrated
probability of correctness.

\subsubsection{CBLP: Class-Balanced Layer Purity}
\label{sec:cblp}

CBLP jointly measures requested-category coverage and cross-category
isolation in delivered layers. Unlike image-level metrics, it distinguishes
separate object layers from mixed-category layers without requiring fixed
layer counts or ground-truth matching.

For scene $i$, \texttt{Qwen3-VL-8B-Instruct} extracts and audits a normalized
standalone target-category set $\mathcal C_i$ from the original prompt, with
$K_i=|\mathcal C_i|$. This set is frozen before inspecting generated outputs.
The same frozen judge detects a set $\mathcal D_{i\ell}$ of distinct standalone
categories in each of the $L_i$ delivered layers, including background and
text layers; RGBA assets are displayed on a uniform gray background.
Detected sets include requested and additional categories under the same
normalization rules. Let $n_{i\ell}=|\mathcal D_{i\ell}|$ be the number of
categories in layer $\ell$.

A missing target category receives zero; otherwise, its score is the
reciprocal of the largest category count among layers containing it:
\begin{equation}
    s_{ic}=\begin{cases}
        0, & c\notin\mathcal D_{i\ell}\ \text{for all }\ell,\\[1mm]
        \displaystyle\frac{1}{\max_{\ell:\,c\in\mathcal D_{i\ell}}n_{i\ell}},
           & \text{otherwise}.
    \end{cases}
    \label{eq:cblp_class}
\end{equation}

Scoring each category by its \emph{most mixed layer} prevents added pure
copies from masking cross-category mixing. A target appearing only in
single-category layers scores $1$.

For scenes with applicable categories,
$\mathcal S=\{i:K_i>0\}$, we define
\begin{equation}
    \mathrm{CBLP}=\frac{1}{|\mathcal S|}\sum_{i\in\mathcal S}
       \left(\frac{1}{K_i}\sum_{c\in\mathcal C_i}s_{ic}\right).
    \label{eq:cblp_merged}
\end{equation}

This gives equal weight to target categories within each scene and to
applicable scenes overall. Scenes with $K_i=0$ are reported as N/A and excluded.

For a cat-and-dog prompt, separate pure layers score $1$, while a single
cat--dog layer or only one pure layer scores $1/2$. Adding pure copies
cannot erase mixing penalties, and extra-only layers do not directly affect
the score. Both AtomicQA and CBLP range from $0$ to $1$ and are multiplied
by $100$ in result tables.

\section{Experiments}
\label{sec:experiments}
We evaluate the quality of layered scenes, the roles of
Planner and object-editor adaptation, the effect of execution-grounded
repair, and the utility of cross-domain training for scene planning.
We assess successive edits via instruction success, content consistency, and qualitative comparisons.

\subsection{Experimental Setup}
\label{sec:exp_setup}
\noindent\textbf{Data and protocol.}
For scene generation and automatic repair, we use
5,000 LayerFolio scene requests. They are sampled proportionally across the
dataset's categories and are disjoint from the 95,377 training scenes
throughout all training stages. All configurations are evaluated against
the same requests; layered methods export RGBA assets.

\noindent\textbf{Inference and evaluation.}
The generated images have a maximum long side of 1,024 pixels.
Repair uses at most three rounds and three candidates; each candidate
allows up to three Editing Agent invocations and two read-only tool
calls, with early termination upon verification. We report CLIP Score~\cite{radford2021clip}
and FID~\cite{heusel2017fid}, together
with AtomicQA and CBLP defined in \S~\ref{sec:requirement_layer_evaluation}. Importantly, the training
reward judge, inference-time Visual Verifier, and frozen final evaluators
are independent.

\subsection{Comparison with Existing Approaches}
\label{sec:exp_comparison}
We compare direct layered generation against LayerDiffuse~\cite{layerdiffuse}
and ART+~\cite{prismlayers}. For context, we also report flat-image
references (Stable Diffusion 3.5 Large~\cite{sd35large} and
Qwen-Image~\cite{qwenimage}) and generation followed by layer
decomposition (Qwen-Image + Qwen-Image-Layered~\cite{qwen_image_layered}).
The latter is evaluated on its exported decomposed results.

\begin{table*}[htbp]
\centering
\small
\setlength{\tabcolsep}{3.0pt}
\renewcommand{\arraystretch}{1.08}
\caption{Comparison of image generation and layered-image delivery methods. Bold indicates the best result among direct layered methods, our primary like-for-like comparison. CBLP is not applicable to flat images; decomposition methods are evaluated on exported layers.}
\label{tab:exp_main}
\begin{tabular}{llccccc}
\toprule
\multirow[c]{2}{*}{\textbf{Route}}
 & \multirow[c]{2}{*}{\textbf{Method}}
 & \textbf{CLIP}
 & \multirow[c]{2}{*}{\textbf{FID}$\,\downarrow$}
 & \multicolumn{2}{c}{\textbf{AtomicQA}}
 & \multirow[c]{2}{*}{\textbf{CBLP}$\,\uparrow$} \\
 & & \textbf{Score}$\,\uparrow$ & & \textbf{AM}$\,\uparrow$ & \textbf{GM}$\,\uparrow$ & \\
\midrule
\multirow[c]{2}{*}{Flat image}
 & Stable Diffusion 3.5 Large~\cite{sd35large} & 40.45 & 26.54 & 71.05 & 63.33 & -- \\
 & Qwen-Image~\cite{qwenimage} & 41.59 & 22.85 & 81.29 & 77.70 & -- \\
\midrule
Gen. + Decomp.
 & Qwen-Image + Qwen-Image-Layered~\cite{qwen_image_layered}
 & 41.57 & 24.35 & 81.15 & 77.62 & 41.57 \\
\midrule
\multirow[c]{3}{*}{Direct layered}
 & LayerDiffuse~\cite{layerdiffuse} & 29.29 & 41.92 & 44.93 & 34.99 & 23.34 \\
 & ART+~\cite{prismlayers} & 38.88 & 27.48 & 65.84 & 56.64 & 43.24 \\
\rowcolor{mygray}
 & FolioWeaver (Ours) & \textbf{40.38} & \textbf{21.24}
 & \textbf{75.64} & \textbf{70.37} & \textbf{54.06} \\
\bottomrule
\end{tabular}
\par\vspace{-6pt} 
\end{table*}

\begin{figure*}[htbp]
\centering
\includegraphics[width=0.95\textwidth]{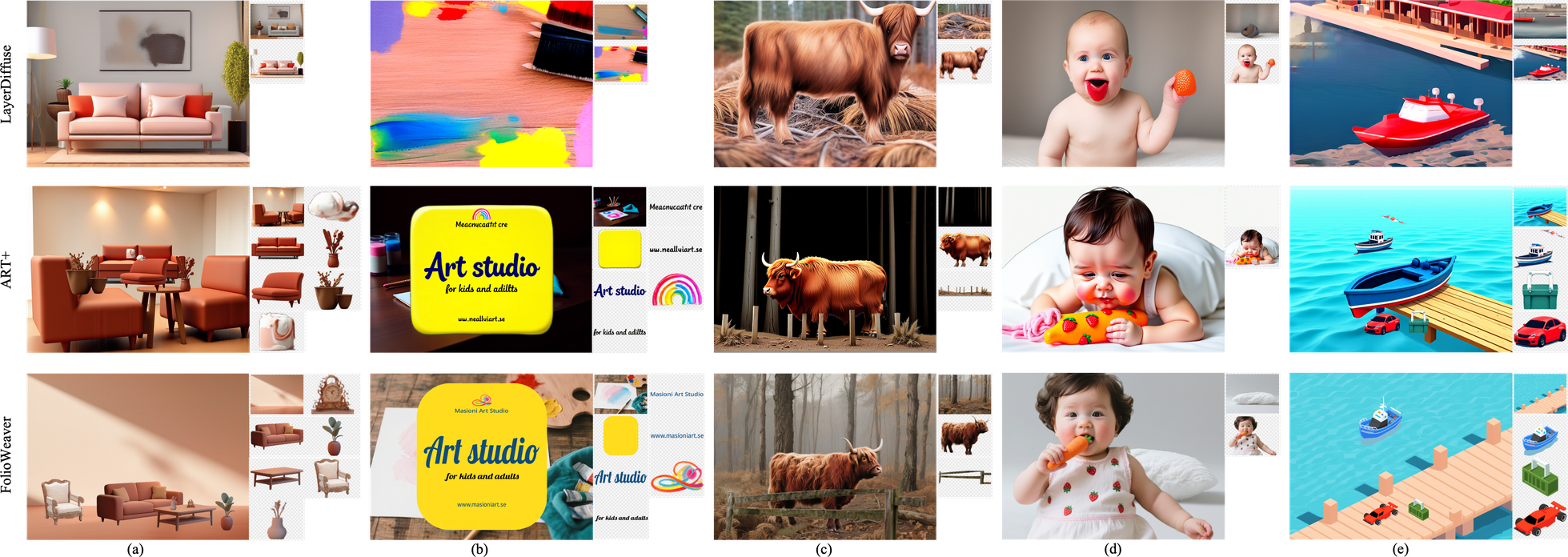}
\caption{Qualitative comparison of LayerDiffuse, ART+, and FolioWeaver: composites and exported layers for matched prompts.}
\label{fig:exp_qual}
\par\vspace{-6pt} 
\end{figure*}

\noindent\textbf{Layer quality and editability.}
As shown in Table~\ref{tab:exp_main}, among methods
that directly generate layers, FolioWeaver improves CBLP
by 10.82 points over ART+ while also improving both AtomicQA scores.
The larger gain in AtomicQA-GM than in AM suggests better support for
weakly satisfied requirements, rather than only a stronger average.
FolioWeaver also attains the lowest FID across the reported systems,
indicating that the layer structure need not come at the
expense of composite-image fidelity. In contrast, although Qwen-Image and its decomposed outputs achieve higher AtomicQA scores, the decomposed layers exhibit substantially lower CBLP, suggesting that strong composite-level performance does not necessarily translate into high-quality layer organization. In
Figure~\ref{fig:exp_qual}, matched interiors, graphic designs, animals,
portraits, and waterfront scenes expose both the composed images and
representative exported assets, making this output distinction visible.

\begin{figure*}[htbp]
\centering
\includegraphics[width=0.95\textwidth]{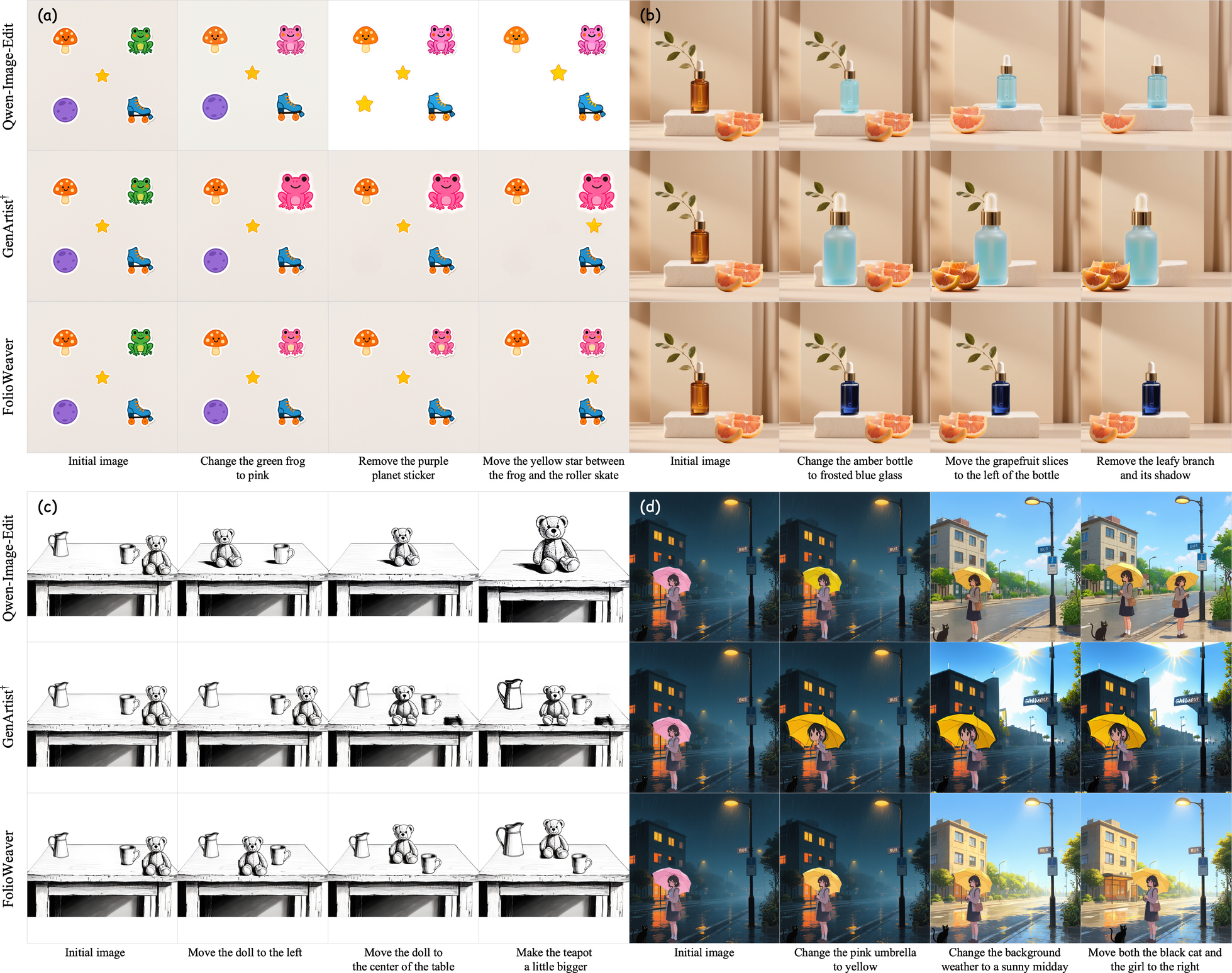}
\caption{Qualitative multi-turn editing comparison: Qwen-Image-Edit, GenArtist$^\dagger$, and FolioWeaver.}
\label{fig:exp_scope}
\par\vspace{-6pt} 
\end{figure*}

\subsection{Component and Agent Policy Ablations}
\label{sec:exp_ablation}
\noindent\textbf{Planner and local object editor.}
We cross native and SFT-adapted Planner and Local Object Editor
checkpoints, keeping the remaining generation pipeline fixed and
evaluating \emph{before repair} (Table~\ref{tab:exp_planner}).
Planner adaptation increases the schema-valid LayerGraph rate by
53.32 percentage points and reduces FID under either editor.
Editor adaptation improves all five visual metrics under either Planner,
with CBLP gains of 3.72 points for the native Planner and 2.92 points for
the adapted Planner. Its FID gain is larger with the native Planner,
whereas its AtomicQA-AM gain is larger with the adapted Planner.
Joint adaptation achieves the best results across all five metrics,
supporting complementary benefits from executable planning and local
asset generation.

\begin{table}[htbp]
\centering\small
\setlength{\tabcolsep}{2.0pt}
\renewcommand{\arraystretch}{1.08}
\caption{Planner \& Editor ablation before repair.
Valid is the percentage of schema-conforming LayerGraph outputs.}
\label{tab:exp_planner}
\begin{tabular}{llcccccc}
\toprule
\multirow[c]{2}{*}{\textbf{Planner}}
 & \multirow[c]{2}{*}{\textbf{Editor}}
 & \multirow[c]{2}{*}{\textbf{Valid}$\,\uparrow$}
 & \textbf{CLIP}
 & \multirow[c]{2}{*}{\textbf{FID}$\,\downarrow$}
 & \multicolumn{2}{c}{\textbf{AtomicQA}}
 & \multirow[c]{2}{*}{\textbf{CBLP}$\,\uparrow$}\\
 & & & \textbf{Score}$\,\uparrow$ & & \textbf{AM}$\,\uparrow$ & \textbf{GM}$\,\uparrow$ & \\
\midrule
Native & Native & 43.54 & 38.74 & 48.82 & 66.14 & 62.55 & 42.15\\
Native & SFT    & 43.54 & 39.38 & 40.90 & 67.67 & 64.35 & 45.87\\
SFT    & Native & \textbf{96.86} & 39.16 & 33.40 & 69.04 & 64.53 & 46.95\\
\rowcolor{mygray}
SFT    & SFT    & \textbf{96.86} & \textbf{39.90} & \textbf{28.61}
               & \textbf{71.95} & \textbf{65.02} & \textbf{49.87}\\
\bottomrule
\end{tabular}
\par\vspace{-6pt} 
\end{table}

\noindent\textbf{Repair policy learning and tool use.}
All repair variants start from identical initial layered
outputs and use the same repair budgets (Table~\ref{tab:exp_agent}).
SFT improves all five metrics over the native Editing Agent.
Without inspection, adding GRPO reduces FID by 2.62 and increases
AtomicQA-GM and CBLP by 2.04 and 1.29 points, respectively, with nearly
unchanged CLIP Score. Enabling inspection further reduces FID by 1.59
and improves both AtomicQA scores and CBLP. Relative to no repair, the
full system improves all five metrics, including a 7.37 reduction in FID
and a 4.19-point gain in CBLP, demonstrating complementary benefits
from policy learning and visual inspection.

\begin{table}[htbp]
\centering\small
\setlength{\tabcolsep}{2.2pt}
\renewcommand{\arraystretch}{1.08}
\caption{Editing Agent ablation with shared initial outputs and budgets.
On/Off denotes read-only tool access.}
\label{tab:exp_agent}
\begin{tabular}{lcccccc}
\toprule
\multirow[c]{2}{*}{\textbf{Policy}}
 & \multirow[c]{2}{*}{\textbf{Tools}}
 & \textbf{CLIP}
 & \multirow[c]{2}{*}{\textbf{FID}$\,\downarrow$}
 & \multicolumn{2}{c}{\textbf{AtomicQA}}
 & \multirow[c]{2}{*}{\textbf{CBLP}$\,\uparrow$}\\
 & & \textbf{Score}$\,\uparrow$ & & \textbf{AM}$\,\uparrow$ & \textbf{GM}$\,\uparrow$ & \\
\midrule
No repair & -- & 39.90 & 28.61 & 71.95 & 65.02 & 49.87\\
Native & Off & 40.11 & 27.47 & 72.33 & 65.95 & 50.40\\
SFT & Off & 40.33 & 25.45 & 74.57 & 67.28 & 51.74\\
SFT+GRPO & Off & 40.32 & 22.83 & 75.58 & 69.32 & 53.03\\
\rowcolor{mygray}
SFT+GRPO & On & \textbf{40.38} & \textbf{21.24} & \textbf{75.64}
                   & \textbf{70.37} & \textbf{54.06}\\
\bottomrule
\end{tabular}
\par\vspace{-6pt} 
\end{table}

\subsection{Cross-Domain Data Utility}
\label{sec:exp_crossdomain}
We train the Planner on Graphic-only (20K), size-matched
Multi-domain (20K), and full-data (95,377) subsets. The two 20K settings
share identical training budgets, with the downstream generator fixed
and repair disabled. In Table~\ref{tab:exp_domains}, Multi-domain improves
all five metrics over Graphic-only, reducing FID by 4.73 and increasing
CBLP by 3.81 points. This matched comparison supports the value of
cross-domain supervision at a fixed training-set size. Full-data further
reduces FID by 1.92 and raises CBLP by 2.66 points, although this
comparison also increases training scale to 95,377 scenes.

\begin{table}[htbp]
\centering\small
\setlength{\tabcolsep}{1.7pt}
\renewcommand{\arraystretch}{1.08}
\caption{Cross-domain Planner training before repair.
Both 20K settings use matched budgets;
Full-data uses more training data.}
\label{tab:exp_domains}
\begin{tabular}{lcccccc}
\toprule
\textbf{Training}
 & \multirow[c]{2}{*}{\textbf{\# Scenes}}
 & \textbf{CLIP}
 & \multirow[c]{2}{*}{\textbf{FID}$\,\downarrow$}
 & \multicolumn{2}{c}{\textbf{AtomicQA}}
 & \multirow[c]{2}{*}{\textbf{CBLP}$\,\uparrow$}\\
\textbf{setting} & & \textbf{Score}$\,\uparrow$ & & \textbf{AM}$\,\uparrow$ & \textbf{GM}$\,\uparrow$ & \\
\midrule
Graphic-only & 20,000 & 39.17 & 35.26 & 67.14 & 63.44 & 43.40\\
Multi-domain & 20,000 & 39.82 & 30.53 & 69.45 & 63.73 & 47.21\\
\rowcolor{mygray}
Full-data & 95,377 & \textbf{39.90} & \textbf{28.61}
          & \textbf{71.95} & \textbf{65.02} & \textbf{49.87}\\
\bottomrule
\end{tabular}
\par\vspace{-6pt} 
\end{table}

\subsection{Multi-Turn Editing Comparison}
\label{sec:exp_feedback}
\noindent
From identical images and instructions, each method edits its own
preceding output. Table~\ref{tab:exp_multiturn} reports turn-averaged
success rate (SR) and consistency score (CS) from a shared frozen
evaluator. FolioWeaver outperforms
both baselines, exceeding GenArtist$^\dagger$ by 8.38 percentage points
in SR and 0.46 in CS. Figure~\ref{fig:exp_scope} illustrates how persistent
layers enable targeted edits while preserving scene content and prior
edits.

\begin{table}[htbp]
\centering\small
\setlength{\tabcolsep}{8pt}
\renewcommand{\arraystretch}{1.08}
\caption{Quantitative multi-turn editing.
$^\dagger$\,Local GenArtist adaptation with model substitutions.}
\label{tab:exp_multiturn}
\begin{tabular}{lcc}
\toprule
\textbf{Method}
 & \textbf{SR}$\,\uparrow$ & \textbf{CS}$\,\uparrow$ \\
\midrule
Qwen-Image-Edit~\cite{qwenimage} & 66.58 & 6.11 \\
GenArtist$^\dagger$~\cite{genartist} & 71.54 & 7.86\\ 
\rowcolor{mygray}
FolioWeaver (Ours) & \textbf{79.92} & \textbf{8.32} \\
\bottomrule
\end{tabular}
\par\vspace{-6pt} 
\end{table}

\section{Conclusion}
\label{sec:con}

We presented \textbf{FolioWeaver}, a LayerGraph-grounded agent that unifies layered-image generation and feedback-driven editing through a persistent Executable Scene State. By combining scoped execution with visual verification, FolioWeaver enables targeted edits while reusing unaffected assets. We also constructed and publicly released \textbf{LayerFolio}, a cross-domain layered dataset, and complementary \textbf{AtomicQA} and \textbf{CBLP} evaluation protocols. Experiments demonstrate improvements over direct layered-generation baselines and highlight the benefits of component adaptation, execution-grounded repair, and cross-domain supervision. These findings establish persistent layer representations as a promising foundation for iterative visual creation.

{\small
\bibliographystyle{ieeenat_fullname}
\bibliography{main}
}

\end{document}